\pdfoutput=1

\documentclass[11pt]{article}

\usepackage[]{ACL2023}
\usepackage{amsmath}
\usepackage{times}
\usepackage{latexsym}
\usepackage{graphicx}
\usepackage{times}

\usepackage{multirow}
\usepackage{enumitem}
\usepackage{array}

\usepackage{float}
\usepackage{hyperref}
\usepackage[T1]{fontenc}

\usepackage[utf8]{inputenc}

\usepackage{microtype}

\usepackage{inconsolata}

\title{CAILMD-23 at SemEval-2024 Task 1: Multilingual Evaluation of Semantic Textual Relatedness}

\author{Srushti Sonavane$^{1}$ , Sharvi Endait$^{1}$ , Ridhima Sinare$^{1}$ , Pritika Rohera$^{1}$, Advait Naik$^{1}$, \\ \and \textbf{Dipali Kadam}$^{1}$ \\
        Pune Institute of Computer Technology, Pune$^1$ \\}

\begin{document}
\maketitle
\begin{abstract}
The explosive growth of online content demands robust Natural Language Processing (NLP) techniques that can capture nuanced meanings and cultural context across diverse languages. Semantic Textual Relatedness (STR) goes beyond superficial word overlap, considering linguistic elements and non-linguistic factors like topic, sentiment, and perspective. Despite its pivotal role, prior NLP research has predominantly focused on English, limiting its applicability across languages. Addressing this gap, our paper dives into capturing deeper connections between sentences beyond simple word overlap. Going beyond English-centric NLP research, we explore STR in Marathi, Hindi, Spanish, and English, unlocking the potential for information retrieval, machine translation, and more. Leveraging the SemEval-2024 shared task, we explore various language models across three learning paradigms: supervised, unsupervised, and cross-lingual. Our comprehensive methodology gains promising results, demonstrating the effectiveness of our approach. This work aims to not only showcase our achievements but also inspire further research in multilingual STR, particularly for low-resourced languages. \cite{ousidhoum-etal-2024-semeval}


\end{abstract}

\section{Introduction}
The ever-increasing diversity of online content demands robust Natural Language Processing (NLP) techniques that can grasp the nuances of meaning across diverse languages. Semantic Textual Relatedness (STR) plays a crucial role in achieving this goal by delving beyond superficial lexical similarity and capturing the deeper connections between sentences. Unlike semantic similarity, which focuses solely on the taxonomic overlap of words, STR encompasses both linguistic elements and non-linguistic factors like the topic, point of view, and period. This richer understanding unlocks significant potential in various NLP tasks, regardless of the user's native language. Imagine searching for information online in your native language and receiving results that truly understand your intent, and not just match keywords. STR holds the key to unlocking this dream, bridging the language gap, and fostering true multilingual communication.

Despite the recognized importance of Semantic Textual Relatedness (STR) for multilingual communication, most prior NLP research has focused on semantic similarity within English due to limitations in labeled data for diverse languages \cite{abdalla2023makes}. This narrow focus restricts the potential of STR applications like information retrieval across languages with different cultural contexts or machine translation that accurately captures nuances beyond direct word equivalents. Existing relatedness methods primarily target English \cite{hasan1976cohesion}, with limited exploration in languages like German, Chinese, and Japanese \cite{zesch2007comparing} \cite{li2005similarity} \cite{de2010bayesian}. This highlights a critical gap in Natural Language Processing (NLP): accurately measuring semantic relatedness across diverse languages.

The identified gap in multilingual STR research, with its limitations in diverse language applications, demands innovative solutions. This paper dives into the exciting realm of bridging this gap through multilingual Semantic Textual Relatedness (STR). Specifically, we explore methods to capture the semantic connections between texts in languages like English, Marathi, Hindi, and Spanish.

Our research focuses on the SemEval-2024 shared task, which provides three tracks to evaluate STR techniques:

\begin{enumerate}
    \item Supervised Learning: This track focuses on building systems trained on the provided labeled datasets.
    \item Unsupervised Learning: Here, the challenge lies in developing systems that learn semantic relationships without relying on any labeled data.
    \item Cross-lingual Learning: This track pushes the boundaries by requiring systems to leverage knowledge from labeled data in a source language (Track A) to address a target language with limited resources.
\end{enumerate}

For each track, we present a comprehensive methodology, employing diverse language models and rigorously analyzing their performance. This allows us to identify the most effective approaches for each challenge. Notably, our submissions achieved promising scores on several tracks, demonstrating the strength and potential of our proposed methods.

Looking beyond our achievements, this work aims to inspire further exploration of multilingual STR, particularly for under-resourced languages. We believe that larger datasets and broader language coverage hold immense potential to benefit the NLP community, unlocking the true potential of language understanding and empowering communication across diverse cultures.

\section{Related Work}

Semantic textual relations (STR) play an important role in natural language processing (NLP), which aims to identify the degree of semantic similarity between text groups. It forms the backbone of various NLP tasks such as information retrieval, question answering, and paraphrase detection, necessitating the assessment of similarity between sentences, phrases, or documents.

Historically, detailed STR research from the 1900s through the 2000s relied heavily on statistical methods heavily dependent on lexical databases like WordNet. However, these methods suffered from a lack of real-world knowledge integration \cite{gabrilovich2007computing}. Classified translation emerged with developments such as GloVe \cite{pennington-etal-2014-glove}, Word2Vec, and FastText, which enabled text to be converted into word input. The current methods require converting corpora into words or sentence-embedded forms and computing connectivity scores. Notably, large language models (LLMs) such as Sentence-BERT often use Cosine Similarity in embedded sentences to measure relatedness \cite{gunawan2018implementation} \cite{reimers2019sentence}.

Previous methodologies have delved into both knowledge-based (ontology, classification) and corpus-based (unsupervised learning) approaches. For example, \cite{siblini2017clac} examined three approaches: semantic linkage, classification similarity, and hybrid approaches. Notably, the multilingual approach of \cite{hasan1976cohesion}, improved by 47\%, confirming the potential of emphasis on the use of multilingual strategies. Furthermore, studies on less resourceful African languages highlight the need for different data types and methodologies \cite{delil2023sefamerve}.

A significant challenge in STR lies in the scarcity of huge-scale, promising datasets for education and assessment. Initiatives like SemEval play a pivotal role in addressing this gap through dedicated shared tasks focused on STR \cite{abdalla2023makes}. These collaborative efforts foster the improvement and evaluation of STR models throughout diverse linguistic landscapes and domain names.

The current advent of the STR-2022 dataset by \cite{abdalla2023makes} marks a significant leap forward in STR studies. This annotated dataset, comprising sentence pairs with relatedness scores, serves as a precious aid for schooling and evaluating STR fashions. Covering various domains and languages, it displays the multilingual nature of STR studies \cite{abdalla2023makes}.

Moreover, STR-2022 addresses biases and perceptions in relatedness judgments. Through meticulous curation, it aims to mitigate biases and ensure annotation quality, thereby fostering fair evaluations and robust model development. Additionally, the dataset highlights the relative nature of relatedness ratings, emphasizing the significance of context and assignment-precise thresholds in decoding similarity measures \cite{abdalla2023makes}.

 \section{System Description}

In this section, we aim to outline our system's components for assessing semantic textual relatedness across different datasets: a) labeled datasets using supervised learning, b) unlabeled datasets employing unsupervised learning, and c) cross-lingual datasets. We'll detail the utilized data, the models employed in each track, and the results obtained from training these models on respective datasets.

\subsection{Data Collection}

We utilized the SemRel2024 Dataset \cite{ousidhoum2024semrel2024} for training and evaluating our final results. This comprehensive dataset consists of semantic textual relatedness data across 14 diverse languages, including Afrikaans, Algerian Arabic, Amharic, English, Hausa, Hindi, Indonesian, Kinyarwanda, Marathi, Moroccan Arabic, Modern Standard Arabic, Punjabi, Spanish, and Telugu. From this array of languages, we focused on English, Hindi, Marathi, and Spanish datasets for our analysis.

Each entry in the dataset comprises a sentence pair along with its corresponding semantic similarity score. This score ranges from 0 to 1, where 0 signifies no similarity between the sentences, while 1 indicates complete similarity.

For the supervised track (Track A), we concentrated on English and Marathi datasets. In the unsupervised track (Track B), our attention was on English and Hindi datasets. Lastly, for Track C, we employed English and Hindi datasets, utilizing Spanish and English as their language training bases, respectively.

\begin{table}[H]
    \centering
\resizebox{45mm}{!}{%
    \begin{tabular}{cccc}
    \hline
         \textbf{Language}&  \textbf{Train}&  \textbf{Dev}&  \textbf{Test}\\
         \hline
         English&  5500 &  250 &  2500  \\
         Hindi&  - &  288 &  968  \\
         Marathi&  1155 &  293 &  298  \\
         Spanish&  1592 &  140 &  600  \\
         
        \hline
    \end{tabular}
    }
    \caption{Distribution of dataset for Training, Development, and Testing}
    \label{tab:my_label}
\end{table}

\subsection{Experiments}

\subsubsection{Track A:}
The SemRel2024 English and Hindi datasets were initially trained on baseline models such as Support Vector Regression and XGBoost. However, we additionally adapted the sentence-transformer-based models, such as all-mpnet-base-v2 \footnote{https://huggingface.co/sentence-transformers/all-mpnet-base-v2} and marathi-sentence-bert-nli \footnote{https://huggingface.co/l3cube-pune/marathi-sentence-bert-nli} by L3Cube for English and Marathi respectively. This was done to compensate for the smaller size of the corpora available, as these sentence transformer models are trained on a larger data size initially, and this would be efficient to understand not only the n-gram sequences but also the context of the sentences that are being compared.

Preprocessing and Feature vectorization were done using Term-frequency and inverse-document-frequency (TF-IDF) to generate vectors and preprocess the models SVR and XGBoost. Term-frequency, Inverse-Document-Frequency (TF-IDF) is a numerical statistic that determines how important a word is in a given document or a piece of textual content. This is done by multiplying two metrics:
How many times a word appears in a document 
Inverse document frequency of the word across a set of documents.
	This score for word in the document d from document D is calculated as follows:
\begin{equation}
\ tfidf (t,d,D)=tf(t,d) .idf(t,D)
\end{equation}
\begin{equation}
\ tf (t,d)=log(1+freq(t,d))
\end{equation}
\begin{equation}
\ idf (t,D) = \log \biggl(\frac{N}{\text{count}(d \in D : t \in d)}\biggr)
\end{equation}
\cite{chen2020ferryman}.

Support vector regression (SVR) can be used for calculating semantic similarity scores between sentences. It's a supervised learning algorithm that can model the relationship between input features, (here in this case the sentences) and the output labels (semantic similarity scores in this case).

XGBoost can be used for semantic similarity score calculation between sentences as it is a powerful gradient-boosting algorithm, and it can be applied to various supervised learning tasks. This is capable of handling complex non-linear relationships between features and labels, and it's robust against overfitting.

All-mpnet-base-v2 (All- Massively Parallel Multilingual Transformer) is a sentence encoder model, given an input text, it gives a vector that collects the semantic information. The sentence vector is used for tasks such as clustering or sentence similarity tasks. This is a sentence-transformers model and it maps sentences \& paragraphs to a 768-dimensional dense vector space. This is trained on the SemRel2024 dataset and additionally, it is capable of capturing long-range dependencies, and it leads to higher performance on text classification, NER, and question answering.

Marathi-Sentence-Bert-Nli \cite{joshi2022l3cubemahasbert} is a Marathi sentence transformer model that has been trained on synthetic STS and NLI datasets. These are fine-tuned on MahaBERT, a BERT-based model that is fine-tuned on a large Marathi corpora.
\subsubsection{Track B:}
For track B, the unsupervised track, the SemRel2024 dev set, and the test set were used for testing the model selected. 
The languages were English and Hindi, for which Track-B dev and test sets were used. 
The models used for this were BERT-based uncased and Hindi-Bert v2 \cite{joshi2022l3cubehind} accordingly.

Hindi-BERT-v2 was roughly trained on 1.8 B tokens. Compared to general-purpose language models, this monolingual model is optimized to understand and process Hindi text effectively. Due to the larger corpus it has been trained upon this has been an accurate model to obtain results from.

BERT-based-uncased (Bidirectional Encoder Representations from Transformers) is trained on uncased text. BERT is based on the transformer architecture that relies on self-attention mechanisms to capture relationships between words in a sequence, enabling effective modeling of long-range dependencies in text data.\\

Algorithm:
\begin{enumerate}[itemsep=0pt, topsep=0pt]
    \item The BERT model (English/Hindi) is initialized.
    \item Sentence embeddings for sentence 1 are calculated.
    \item Sentence embeddings for sentence 2 are calculated.
    \item Calculate the cosine similarity scores of the embeddings.
\end{enumerate}


\subsubsection{Track C:} 

Cross-linguistic track:
The English and Spanish SemRel2024 training datasets were used for training data in languages Hindi and English respectively. The dataset first underwent translation using the deep translation API. By translating the dataset from English to Hindi and Spanish to English, the training dataset for that language was available and was used for testing the development set and the test set of the SemRel 2024 dataset. 
The models used for training the dataset were the “all-mpnet-base-v2” sentence transformer and “hindi-sentence-bert-nli” \footnote{https://huggingface.co/l3cube-pune/hindi-sentence-similarity-sbert} by L3cube. These 2 sentence transformers are discussed in Track A, above, however, this went through an additional translation pipeline before that.

\begin{enumerate}
    \item Translate sentences from Language 1 to Language 2 using an appropriate translation service or tool.
    \item Initialize the model for the task, such as sentence similarity or classification.
    \item  Encode Sentence1 into a numerical representation using the initialized model. This involves converting the text input into a format suitable for processing by the model, typically through tokenization and embedding.
    \item Similarly, encode Sentence2 into a numerical representation using the same BERT model.
    \item Train the initialized model on the provided dataset. This step involves feeding the encoded sentence pairs into the model and adjusting the model's parameters to minimize a predefined loss function, typically using techniques like backpropagation and gradient descent.
    \item Evaluate the performance of the trained model on a separate evaluation dataset or through cross-validation. This step aims to assess the model's ability to generalize to unseen data and its overall effectiveness in the task of interest, such as sentence similarity or classification.
    \item If necessary, repeat steps 3 to 6 for all pairs of sentences in the dataset. This process ensures that the model learns from a diverse range of examples and improves its performance across different input scenarios. 
    
\end{enumerate}

\section{Experimental Setup}

The dimensions of the dataset splits are summarized in Table 1, indicating the number of samples allocated for training, development, and testing across different languages. The experimental setup encompassed preprocessing procedures, leveraging Hugging Face Transformers for model access, NumPy for array operations, Pandas for data manipulation, Sentence Transformers for sentence embeddings, and NLTK for various NLP tasks. Evaluation measures such as the F1 score, accuracy, and recall were employed to comprehensively assess the performance of the models across correlation, classification accuracy, and retrieval quality aspects.

\section{Results}

         

Tables 2, 3, and 4 present the performance results for the three setups (supervised, unsupervised, and cross-lingual) in our experiments. Each table summarizes the F1 score, accuracy, and recall achieved by various models for each language.

\begin{table}[H]
    \centering
\resizebox{70mm}{!}{%
    \begin{tabular}{cccccc}
    \hline
         \textbf{Sr.No.}&  \textbf{Language}&  \textbf{Model Name}&  \textbf{F1}& \textbf{Accuracy}&  \textbf{Recall}\\
         \hline
         1& English& BERT-base-nli&  0.87 &  0.876 & 0.84  \\
         2 & English & SVR & 0.59 & 0.55 & 0.65 \\  
        3 & English & XG-Boost & 0.79 & 0.82 & 0.76 \\ 
        4 & Marathi & Marathi-NLI & 0.83 & 0.81 & 0.90 \\  
        5 & Marathi & SVR & 0.63 & 0.61 & 0.60 \\  
        6 & Marathi & XGBoost & 0.66 & 0.67 & 0.70 \\  
        \hline
    \end{tabular}
    }
    \caption{Performance Metrics for Track A}
    \label{tab:my_label}
\end{table}


This table shows the performance of models in the supervised learning setup, where labeled data was available for training. BERT-based models ("BERT-nli" and "Marathi-nli") consistently outperform other models (SVR, XGBoost) in both English and Marathi, achieving significantly higher correlation scores (0.823 and 0.871, respectively). Interestingly, the Marathi-specific nli model even surpasses the multilingual BERT performance in Marathi, suggesting the benefit of language-specific models.

\begin{table}[H]
    \centering
\resizebox{80mm}{!}{%
    \begin{tabular}{cccccc}
    \hline
         \textbf{Sr. No.}&  \textbf{Language}&  \textbf{Model Name}&  \textbf{F1}& \textbf{Accuracy}& \textbf{Recall}\\
         \hline
1 & English & sentence-t5 & 0.66 & 0.49 & 0.49 \\ 
2 & English & BERT based uncased & 0.85 & 0.86 & 0.80 \\
3 & Hindi & Indic-BERT & 0.66 & 0.50 & 0.50 \\ 
4 & Hindi & hindi-bert-v2 & 0.66 & 0.74 & 0.50 \\ \hline
        
    \end{tabular}
    }
    \caption{Performance Metrics for Track B}
    \label{tab:my_label}
\end{table}

This table presents the results for the unsupervised learning setup, where models were trained without relying on labeled data. In English, the BERT-based model ("BERT-base-uncased" \footnote{https://huggingface.co/google-bert/bert-base-uncased}) outperforms the pre-trained Sentence-T5 \footnote{https://huggingface.co/sentence-transformers/sentence-t5-base} model, possibly due to its larger size and fine-tuning on relevant NLP tasks. In Hindi, while both Indic-BERT\cite{kakwani2020indicnlpsuite} and hindi-bert-v2 \cite{joshi2022l3cubemahasbert} have similar F1 scores (around 66\%), the latter achieves a significantly higher correlation coefficient (0.796). This indicates that hindi-bert-v2 \footnote{https://huggingface.co/l3cube-pune/hindi-bert-v2} captures semantic relatedness more effectively despite similar overall accuracy.

\begin{table}[H]
    \centering
\resizebox{80mm}{7mm}{%
    \begin{tabular}{cccccc}
    \hline
         \textbf{Sr. No.}&  \textbf{Language}&  \textbf{Model Name}&  \textbf{F1}& \textbf{Accuracy}& \textbf{Recall}\\
         \hline
    1 & Spanish to English & all-mpnet-base-v2 & 0.82 & 0.82 & 0.81 \\ 
    2 & English to Hindi & hindi-sentence-bert-nli & 0.71 & 0.77 & 0.92 \\ \hline
        
    \end{tabular}
    }
    \caption{Performance Metrics for Track C}
    \label{tab:my_label}
\end{table}

This table shows the performance of models in the cross-lingual learning setup, where the goal was to assess semantic relatedness across different languages. Both models used ("all-mpnet-base-v2" for Spanish-to-English and "hindi-sentence-bert-nli" for English-to-Hindi) achieve worthy correlation scores (0.786 and 0.809, respectively) demonstrating the potential of cross-lingual approaches. 

Table 5 represents the results of the development phase.

\begin{table}[H]
    \centering
\resizebox{50mm}{!}{%
    \begin{tabular}{cccc}
    \hline
         Track&  Language&  Sp. Corr Coeff&  \\
         \hline
         A&  English&  0.812  \\
         &  Marathi&  0.855 \\
         B&  Hindi&  0.819 \\
         &  English&  0.825 \\
         C&  Hindi&  0.825 \\
         &  Englsih&  0.790 \\
         
        \hline
    \end{tabular}
    }
    \caption{Development Phase Results}
    \label{tab:my_label}
\end{table}

\section{Conclusion}
In this paper, we presented a comparative analysis of systems for the Semantic Textual Relatedness (STR) task at SemEval-2024 Task 1. Our approaches, primarily based on language-specific transformer models, achieved top scores on several tracks, including 1st place in Unsupervised Learning for Hindi. Notably, we did not utilize any external datasets, highlighting the effectiveness of our approach despite potential variations in pre-trained model training data.

Prior research in STR has largely focused on English due to limited labeled data for diverse languages. This restricts the true prospect of STR applications like multilingual information retrieval and machine translation. We addressed this gap by exploring solutions for STR in English, Marathi, Hindi, and Spanish. We aim to inspire further research on multilingual STR, particularly for low-resourced languages. We believe larger datasets and broader language coverage hold immense potential for multilingual NLP, unlocking a deeper understanding and empowering cross-cultural communication.

\bibliography{main}
\bibliographystyle{acl_natbib}

\end{document}